\documentclass[splncs03]{llncs}
\usepackage{llncsdoc}
\usepackage{graphicx}
\usepackage{array}
\usepackage{subfigure}
\usepackage{epstopdf}
 \usepackage{amsmath}
 \usepackage{amsfonts}
 
 \usepackage{algorithm}
 \usepackage{algorithmic}
 \usepackage{multirow}
\begin{document}
 
\newcounter{save}\setcounter{save}{\value{section}}
{\def\addtocontents#1#2{}%
\def\addcontentsline#1#2#3{}%
\def\markboth#1#2{}%
\title{Deep Discrete Hashing with Self-supervised Pairwise Labels}

\author{Jingkuan Song \and  Tao He\and Hangbo Fan \and Lianli Gao\thanks{Corresponding Author}}
\institute{University of Electronic Science and Technology of China\\
	2006th Xiyuan Ave, Chengdu, China\\
	\{jingkuan.song,tao.he,hbfan\}@gmail.com,lianli.gao@uestc.edu.cn}

\maketitle
\begin{abstract}
Hashing methods have been widely used for applications of large-scale image retrieval and classification. 
Non-deep hashing methods using handcrafted features have been significantly outperformed by deep hashing methods due to their better feature representation and end-to-end learning framework. However, the most striking successes in deep hashing have mostly involved discriminative models, which require labels.
In this paper, we propose a novel unsupervised deep hashing method, named Deep Discrete Hashing (DDH), for large-scale image retrieval and classification.
In the proposed framework, we address two main problems: 1) how to directly learn discrete binary codes? 2) how to equip the binary representation with the ability of accurate image retrieval and classification in an unsupervised way? 
We resolve these problems by introducing an intermediate variable and a loss function steering the learning process, which is based on the neighborhood structure in the original space.
Experimental results on standard datasets (CIFAR-10, NUS-WIDE, and Oxford-17) demonstrate that our DDH significantly outperforms existing hashing methods by large margin in terms of~mAP for image retrieval and object recognition. Code is available at \url{https://github.com/htconquer/ddh}.
\end{abstract}
\section{Introduction}
Due to the popularity of capturing devices and the high speed of network transformation,  we are witnessing the explosive growth of images, which attracts great attention in computer vision to facilitate the development of multimedia search \cite{DBLP:conf/ijcai/WanWHZGWZL15,Song:2013:IHL:2463676.2465274}, object segmentation \cite{DBLP:conf/ijcai/LiLWLM15,Song:2016:JGL:2964284.2964295}, object detection \cite{DBLP:conf/ijcai/NguyenS15}, image understanding \cite{gao2015optimal,chen2016sca} etc. 
Without a doubt, the ever growing abundance of images brings an urgent need for more advanced large-scale image retrieval technologies. To date, high-dimensional real-valued features descriptors (e.g., deep Convolutional Neural Networks (CNN) \cite{DBLP:journals/ijcv/RussakovskyDSKS15,DBLP:journals/corr/SzegedyIV16,DBLP:journals/corr/TargAL16} and SIFT descriptors) demonstrate superior discriminability, and bridge the gap between low-level pixels and high-level semantic information. But they are less efficient for large-scale retrieval due to their high dimensionality.

Therefore, it is   necessary to transform these high-dimensional features into compact binary codes which enable machines to run retrieval in real-time and with low memory.
Existing hashing methods can be classified into two categories: \textit{data-independent} and \textit{data-dependent}. For the first category, hash codes are generated by randomly projecting samples into a feature space and then performing binarization, which is independent of any training samples. 
On the other hand, \textit{data-dependent} hashing methods learn hash functions by exploring the distribution of the training data and therefore, they are also called learning to hashing methods (L2H) \cite{DBLP:journals/corr/WangZSSS16}.
A lot of L2H methods have been proposed, such as Spectral hashing (SpeH) \cite{NIPS2008_3383}, iterative quantization (ITQ) \cite{DBLP:journals/pami/GongLGP13}, Multiple Feature Hashing (MFH) \cite{song2013effective}, Quantization-based Hashing (QBH) \cite{Song2017}, K-means Hashing (KMH) \cite{DBLP:conf/cvpr/HeWS13}, DH \cite{7298862}, DPSH \cite{DBLP:conf/ijcai/LiWK16}, DeepBit \cite{Lin_2016_CVPR}, etc. 
Actually, those methods can be further divided into two categories: supervised methods and unsupervised methods. The difference between them is whether to use supervision information, e.g., classification labels. Some representative unsupervised methods include ITQ, Isotropic hashing \cite{NIPS2012_4846}, and DeepBit which achieves promising results, but are usually outperformed by supervised methods.
By contrast, the supervised methods take full advantage of the supervision information. One representative is DPSH \cite{DBLP:conf/ijcai/LiWK16}, which is the first method that can perform simultaneous feature learning and hash codes learning with pairwise labels.
However, the information that can be used for supervision is also typically scarce.

To date, hand-craft floating-point descriptors such as SIFT, Speeded-up Robust Features (SURF)~\cite{Bay2008346}, DAISY \cite{4815264}, Multisupport Region Order-Based Gradient Histogram (MROGH)~\cite{6112779}, the Multisupport Region Rotation and Intensity Monotonic Invariant Descriptor (MRRID) \cite{6112779} etc, are widely utilized to support image retrieval since they are distinctive and invariant to a range of common image transformations. 
In \cite{PanJLZSK16}, they propose a content similarity based fast reference frame selection algorithm for reducing the computational complexity of the multiple reference frames based inter-frame prediction. In \cite{TianC17}, they develop a so-called correlation component manifold space learning (CCMSL) to  learn a common feature space by capturing the correlations between the heterogeneous databases.
Many attempts \cite{BHW10,6619157} were focusing on compacting such high quality floating-point descriptors for reducing computation time and memory usage as well as improving the matching efficiency. In those methods, the floating-point descriptor construction procedure is independent of the hash codes learning and still costs a massive amounts of time-consuming computation. Moreover, such hand-crafted feature may not be optimally compatible with hash codes learning. Therefore, these existing approaches might not achieve satisfactory performance in practice. 

To overcome the limitation of existing hand-crafted feature based methods, some deep feature learning based deep hashing methods \cite{DBLP:conf/ijcai/LiWK16,CNNH,guo2016hash,do2016learning,DBLP:conf/ijcai/LiWK16,yang2017supervised,gu2016supervised} have recently been proposed to perform simultaneous feature learning and hash-code learning with deep neural networks, which have shown better performance than traditional hashing methods with hand-crafted features. Most of these deep hashing methods are supervised whose supervision information is given as triplet or pairwise labels.
An example is the deep supervised hashing method by Li \textit{et al.} \cite{DBLP:conf/ijcai/LiWK16}, which can simultaneously learn features and hash codes. Another example is Supervised Recurrent Hashing (SRH)~\cite{gu2016supervised} for generating hash codes of videos. Cao \textit{et al.} \cite{cao2017hashnet} proposed a continuous method to learn binary codes, which can avoid the relaxation of binary constraints~\cite{gu2016supervised} by first learning continuous representations and then thresholding them to get the hash codes. They also added weight to data for balancing similar and dissimilar pairs.

In the real world, however, the vast majority of training data do not have labels, especially for
scalable dataset. 
To the best of our knowledge, DeepBit \cite{Lin_2016_CVPR} is the first to propose a deep neural network to learn binary descriptors in an unsupervised manner, by enforcing three criteria on binary codes. It achieves the state-of-art performance for image retrieval, but DeepBit does not consider the data distribution in the original image space. Therefore, DeepBit misses a lot of useful unsupervised information.

So can we obtain the pairwise information by exploring the data distribution, and then use this information to guide the learning of hash codes?
Motivated by this, in this paper, we propose a Deep Discrete Hashing (DDH) with pseudo pairwise labels which makes use the self-generated labels of the training data as supervision information to improve the effectiveness of the hash codes. It is worth highlighting the following contributions:
\begin{enumerate}
	\item We propose a general end-to-end learning framework to learn discrete hashing code in an unsupervised way to improve the effectiveness of hashing methods.  The discrete binary codes are directly optimized from the training data. We solve the discrete hash, which is hard to optimize, by  introducing an intermediate variable.
	\item To explore the data distribution of the training images, we learn on the training dataset and generate the pairwise information. We then train our model by using this pairwise information in a self-supervised way.
	\item Experiments on real datasets show that DDH achieves significantly better performance than the state-of-the-art unsupervised deep hashing methods in image retrieval applications and object recognition.
	\end {enumerate}

\section{Our Method}
Given $N$ images, $\textbf{I} = \{\textbf{I}_i\}^N_{i=1}$ without labels, our goal is to learn their compact binary codes $\textbf{B}$ such that: (a) the binary codes can preserve the data distribution in the original space, and (b) the discrete binary codes could be computed directly.

As shown in Fig.~ \ref{framework},
 our DDH consists of two key components: construction of pairwise labels, and hashing loss definition. For training, we first construct the neighborhood structure of images and then train the network. For testing, we obtain the binary codes of an image by taking it as an input. In the remainder of this section, we first describe the process of constructing the neighborhood structure, and then introduce our loss function and the process of learning the parameters.

\subsection{Construction of Pairwise Labels }

  In our unsupervised approach, we propose to exploit the neighborhood structure of the images in a feature space as information source steering the process of hash learning. Specifically, we propose a method based on the K-Nearest Neighbor (KNN) concept to create a neighborhood matrix $\textbf{S}$. Based on \cite{He2015}, we extract 2,048-dimensional features from the pool5-layer, which is last layer of ResNet~\cite{He2015}. This results in the set $\textbf{X} = \{{\textbf{x}_{i}}\}^{N}_{i=1}$ where $\textbf{x}_{i}$ is the feature vector of image~$\textbf{I}_i$.
 \begin{figure} 
 	\centering
    \includegraphics[width=1.0 \textwidth]{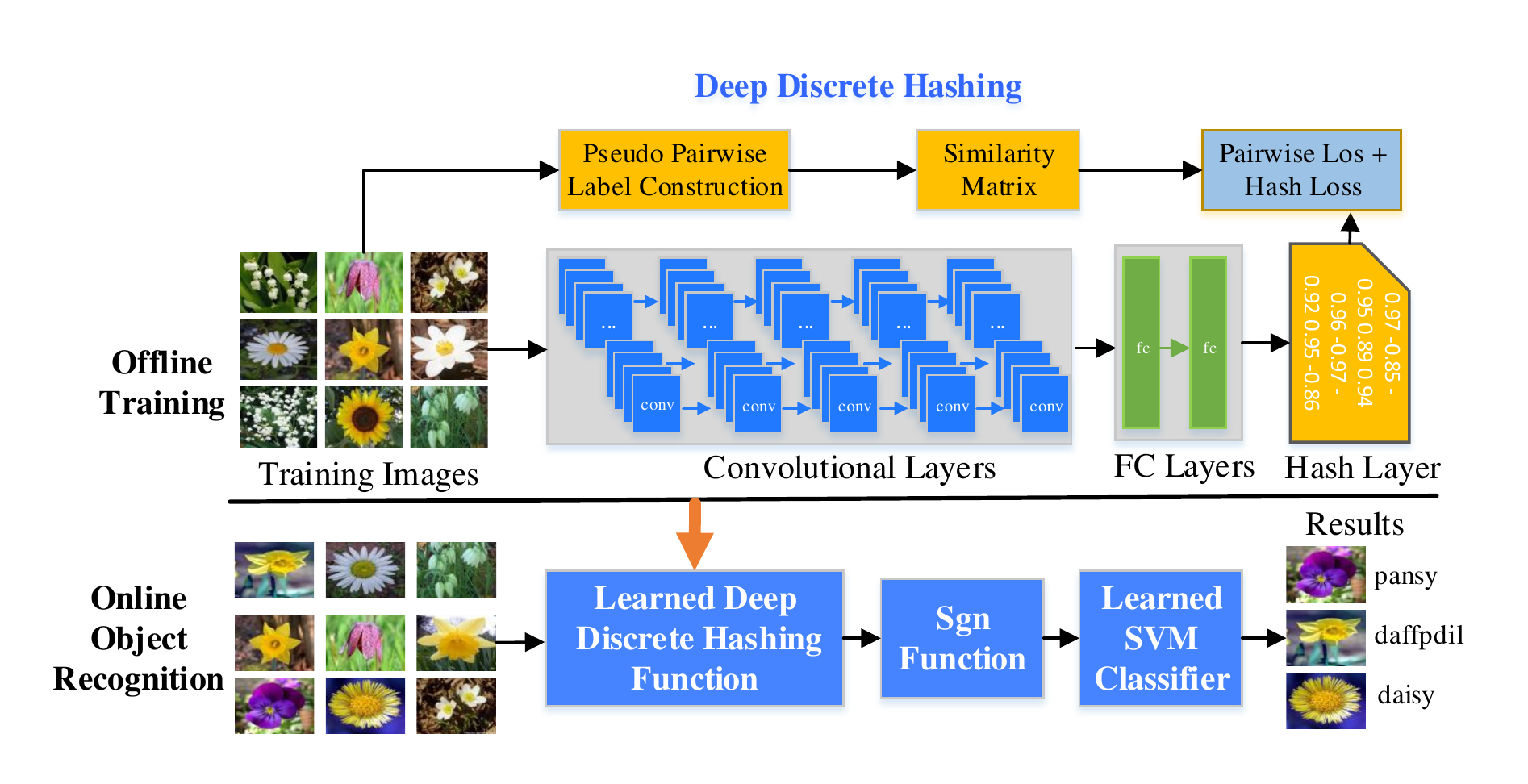}
    \caption{The structure of our end-to-end framework. It has two components, construction of pairwise labels, and hashing loss definition. We first construct the neighborhood structure of images and then train the network based on the define loss function. We utilize the deep neural network to extract the features of the images.}
    \label{framework}
 \end{figure} 
 
 For the representation of the neighboring structure, our task is to construct a matrix $\textbf{\textit{S}}=(s_{ij})^N_{i,j=1}$, whose elements indicate the similarity ($s_{ij}=1$) or dissimilarity ($s_{ij} = -1$) of any two images $i$ and $j$ in terms of their features $\textbf{x}_{i}$ and $\textbf{x}_{j}$.

 We compare images using cosine similarity of the feature vectors. For each image, we select $K_1$ images with the highest cosine similarity as its neighbors. Then we can construct an initial similarity matrix $\textbf{S}_1$:
 \begin{eqnarray}
 {{\left(S_1\right)}_{ij}} = \left\{ {\begin{array}{*{20}{l}}
 	{1,~\textrm{if}~\textbf{x}_{j}~\textrm{is $K_1$-NN of}~\textbf{x}_i}\\
 	{0,~\textrm{otherwise}}
 	\end{array}} \right.
 \end{eqnarray}
 
 Here we use $\textbf{L}_1 ,  \textbf{L}_2  , ..., \textbf{L}_N $ to denote the ranking lists of points $\textbf{I}_1$,$\textbf{I}_2$,...,$\textbf{I}_N$ by \textrm{${K}_1$-NN}. The precision-recall curve in Fig.~\ref{fig1} indicates the quality of the constructed neighborhood for different values of ${K}_1$. Due to the rapidly decreasing precision with the increase of ${K}_1$, creating a large-enough neighborhood by simply increasing ${K}_1$ is not the best option. In order to find a better approach, we borrow the ideas from the domain of graph modeling. In an undirected graph, if a node $v$ is connected to a node $u$ and if $u$ is connected to a node $w$, we can infer that $v$ is also connected to $w$. Inspired by this, if we treat every training image as a node in an undirected graph, we can expand the neighborhood of an image $i$ by exploring the neighbors of its neighbors. Specifically, if $\textbf{x}_{i}$ connects to $\textbf{x}_{j}$ and $\textbf{x}_{j}$ connects to $\textbf{x}_{k}$, we can infer that $\textbf{x}_{i}$ has the potential to be connected to $\textbf{x}_{k}$ as well.
 
 \begin{figure}[h]
 	\centering
 	\includegraphics[width=0.8\linewidth,height=7cm]{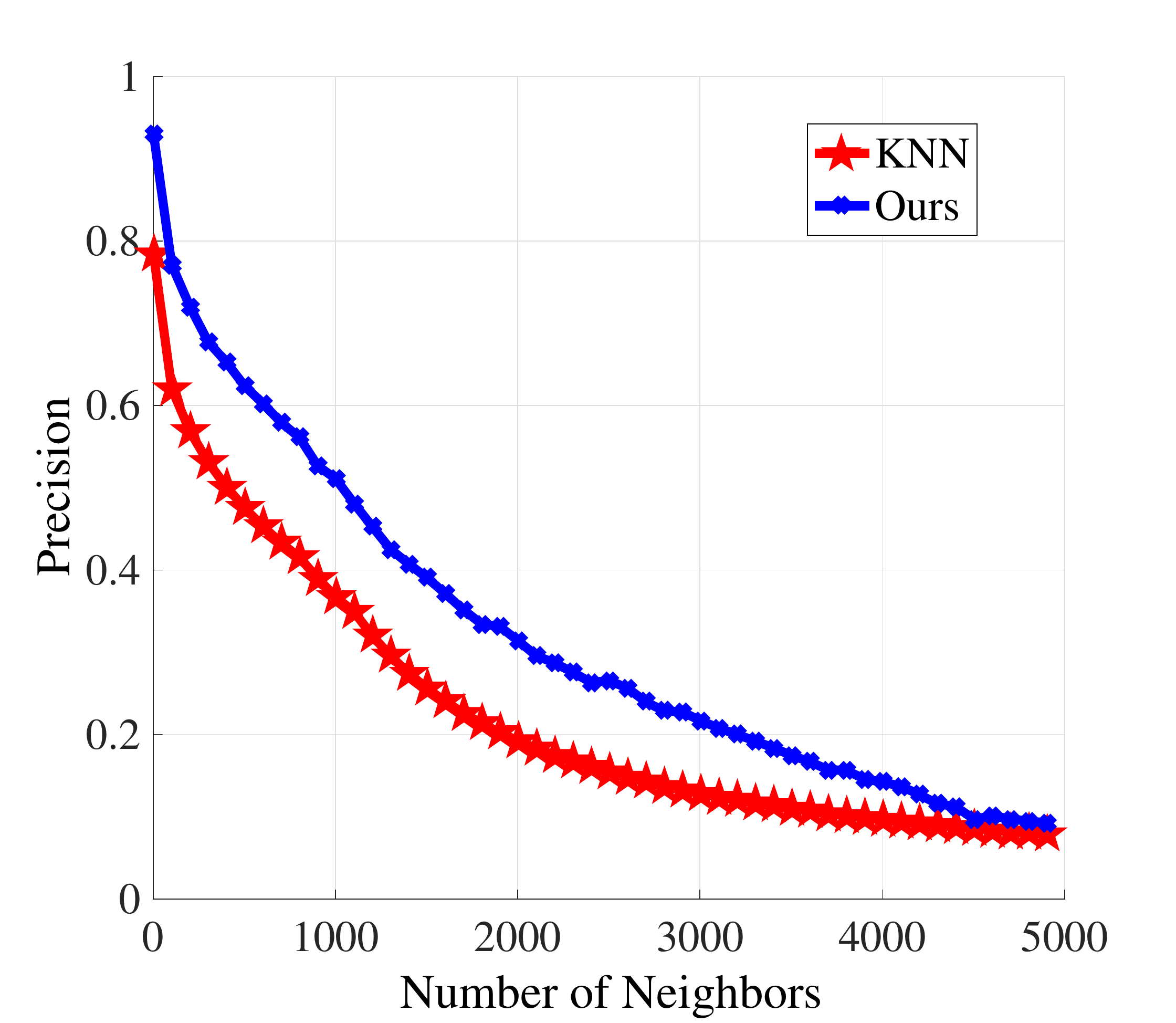}
 	\caption{Precision of constructed labels on cifar-10 dataset with different K, and different methods.}
 	\label{fig1}
 \end{figure}

 Based on the above observations, we construct $\textbf{S}_1$ using the deep CNN features. If we only use the constructed labels by $\textbf{S}_1$, each image has too few positive labels with high  precision. So we increase the number of neighbors based on $\textbf{S}_1$ to obtain more positive labels.  Specifically, we calculate the similarity of two images {by comparing the two ranking lists of ${K}_1$-NN using the expression }{$\frac{1}{||\textbf{L}_i- \textbf{L}_j||^2}$}.   Actually, if two images have the same labels, they should have a lot of intersection points based on $K_1$-NN, i.e., they  have similar ${K}_1$-NN ranking list.  Then we again construct a ranking list of $K_2$ neighbors, based on which we generate the second similarity matrix $\textbf{S}_2$ as:
 \begin{equation}
 {{\left(\textbf{S}_2\right)}_{ij}} = \left\{ {\begin{array}{*{20}{l}}
 	{1,~\textrm{if}~\textbf{L}_{j}~\textrm{is $K_2$-NN of}~\textbf{L}_i}\\
 	{0,~\textrm{otherwise}}
 	\end{array}} \right.
 \end{equation}
 Finally, we construct the neighborhood structure by combining the direct and indirect similarities from the two matrices together. This results in the final similarity matrix $\textbf{S}$:	
 \begin{equation}
{S_{ij}} = \left\{ {\begin{array}{*{20}{l}}
	{1,  ~\textrm{if}~{\left(\textbf{S}_2\right)}_{ik}=1~ and ~ j~ in ~\textbf{L}_k}\\
	{0,~\textrm{otherwise}}
	\end{array}} \right.
\label{eq.neigh.exp}
\end{equation}
where the ${\textbf{L}}_k$ is the ranking list after ${K}_1$-NN.
The whole algorithm is shown in Alg.~\ref{alg.pairwise}. After the two steps KNN, the constructed label precision is shown in Fig.~\ref{fig1}. We note that we could have also omitted this preprocessing step and construct the neighborhood structure directly during the learning of our neural network. We found, however, that the construction of neighborhood structure is time-consuming, and that updating of this structure based on the updating of image features in each epoch does not have significant impact on the performance. Therefore, we chose to obtain this neighborhood structure as described above and fix it for the rest of the process.

\begin{algorithm}[h]
	\begin{algorithmic}[1]
		\renewcommand{\algorithmicrequire}{\textbf{Input:}}
		\renewcommand{\algorithmicensure}{\textbf{Output:}}
		\REQUIRE Images $\textbf{X}  = \{ {\textbf{x}_i}\} _{i = 1}^N$, the number of neighbors $K_1$, the number of neighbors $K_2$ for the neighbors expansion;
		\ENSURE Neighborhood matrix $\textbf{S} = \{ {s_{ij}}\}$;
		\STATE	{First ranking:} Use cosine similarity to generate the index of $K_1$-NN of each image $
		L_1, L_2, ..., L_N$;\\
		\STATE {Neighborhood expansion}:\\
		\FOR {$i$=1,...,$N$}
		\STATE Initialize $num \leftarrow  \emptyset $;\\
		\FOR {$j$=1,...$N$}
		\STATE $num_j \leftarrow$ the size of ${L_i} \cap {L_j}$;\\
		\ENDFOR
		\STATE Sort $num$ by descending order and keep the top $K_2$ $\{L_j\}$; \\
		\STATE Set new ${{L'}_i}$ $ \leftarrow $  union of the top $K_2$ $\{L_j\}$;\\
		\ENDFOR
		\FOR {$j$=1,...,$N$}
		\STATE Construct \textbf{S} with new ${{L'}_j}$ base on Eq.\ref{eq.neigh.exp};\\
		\ENDFOR
		\RETURN $\textbf{S}$;\\
		\caption{Construction of neighborhood structure}
		\label{alg.pairwise}
	\end{algorithmic}
\end{algorithm}

\subsection{Architecture of Our Network}
We introduce an unsupervised deep  framework, dubbed  Deep Discrete Hashing (DDH), to learn compact yet discriminative binary descriptors. The framework includes two main modules, feature learning part and hash codes learning part, as shown in Figure \ref{framework}. 
More specifically, for the feature learning, we use a similar network as in \cite{Zhang:2014:SHL:2600428.2609600}, which has seven layers and the details are shown in Table~\ref{table1}.
In the experiment, we can easily replace the CNN-F network with other deep networks such as \cite{krizhevsky2012imagenet}, \cite{GoogleNet} and \cite{He2015}. Our framework has two branches with the shared weights and both of them have the same weights and same network structure.

We discard the last softmax layer and replace it with a hashing layer, which consists of a fully connected layer and a sgn activation layer to generate compact codes. Specifically, the output of the $full_7$ is firstly mapped to a $L$-dimensional real-value code, and then a binary hash code is learned directly, by converting the $L$-dimensional representation to a binary hash code $\textbf{b}$ taking values of either $+1$ or $-1$. This binarization process can only be performed by taking the sign function $\textbf{b} = sgn(\textbf{u})$ as the activation function on top of the hash layer.
\begin{equation}
\textbf{b} = {\mathop{\rm sgn}} (\textbf{u}) = \left\{ {\begin{array}{*{20}{l}}
	{ + 1,~if~\textbf{u} \ge 0}\\
	{ - 1,~~\textrm{otherwise}}
	\end{array}} \right.
\label{eq.sgn}
\end{equation}

\begin{table}[h]
	\centering
	\caption{The configuration of our framework}
	\label{table1}
	\begin{tabular}{ l cccc }
		\hline\noalign{\smallskip}
		Layer & Configure      \\
		\noalign{\smallskip}
		\hline
		\noalign{\smallskip}
		
		$conv_1$ & filter 64x11x11, stride 4x4, pad 0, LRN, pool 2x2 \\  
		$conv_2$ & filter 256x5x5, stride 1x1, pad 2, LRN, pool 2x2  \\  
		$conv_3$ & filter 256x3x3, stride 1x1, pad 1       \\  
		$conv_4$ & filter 256x3x3, stride 1x1, pad 1         \\  
		$conv_5$ & filter 256x3x3, stride 1x1, pad 1, pool 2x2         \\  
		$full_6$ & 4096                                 \\ 
		$full_7$ & 4096                                 \\ 
		$hash~layer$ & $L$                                 \\ 
		\hline
		
	\end{tabular}
\end{table}

\subsection{Objective Function}

Suppose we denote the binary codes as $\textbf{B} = \left\{ {{\textbf{b}_i}} \right\}_{i = 1}^N$ for all the images. 
The neighborhood structure loss models the loss in the similarity structure in data, as revealed in the set of neighbors obtained for an image by applying the hash code of that image.
We define the loss function as below:
\begin{equation}
\label{Eq.1}
\min {J_1} = \frac{1}{2}{\sum\limits_{{s_{ij}} \in S} {\left( {\frac{1}{L}{\textbf{b}_i}^T{\textbf{b}_j} - {s_{ij}}} \right)} ^2}
\end{equation}
where $L$ is the length of hashing code  and ${s_{ij}} \in \{  - 1,1\}$ indicates the similarity of image $i$ and $j$.
The goal of optimizing for this loss function is clearly to bring the binary codes of similar images as close to each other as possible.

We also want to minimize the quantization loss between the learned binary vectors $\textbf{B}$ and the original real-valued vectors $\textbf{Z}$. It is defined as:
\begin{equation}
\label{Eq.quan}
\min {J_2} = \frac{1}{2}\sum\limits_{i = 1}^N {{{\left\| {{\textbf{z}_i} - {\textbf{b}_i}} \right\|}^2}}
\end{equation}
where $\textbf{z}_i$ and $\textbf{b}_i$ are the real-valued representation and binary codes of the $i$-th image in the hashing layer. Then we can obtain our final objective function as:
\begin{equation}
\label{Eq.obj1}
\min {J} = J_1+\lambda_1J_2, \qquad{{\textbf{b}}_{i}}\in {{\{-1,1\}}^{L}},~~\forall i=1,2,3,...N
\end{equation}
where $\lambda_1$ is the parameter to balance these two terms.

Obviously, the problem in (\ref{Eq.obj1}) is a discrete optimization problem, which is hard to solve. 
LFH \cite{Zhang:2014:SHL:2600428.2609600} solves it by directly relaxing $\textbf{b}_{i}$ from discrete to continuous, which might not achieve satisfactory performance \cite{kang2016column}.
In this paper, we design a novel strategy which can solve the problem \ref{Eq.1} by introducing an intermediate variable. First, we reformulate the problem in \ref{Eq.1} as the following equivalent one:
\begin{equation}
\label{equ2}
\begin{aligned}
& \min {{J}}=\frac{1}{2}\sum\limits_{{{s}_{ij}}\in S}{{{\left( \frac{1}{L}{{\textbf{b}}_{i}}^{T}{{\textbf{b}}_{j}}-{{s}_{ij}} \right)}^{2}}} + \frac{\lambda_1}{2}\sum\limits_{i = 1}^N {{{\left\| {{\textbf{z}_i} - {\textbf{b}_i}} \right\|}^2}} \\ 
& s.t \quad{{\textbf{u}}_{i}}={{\textbf{b}}_{i}},\forall i=1,2,3,...N \\ 
& \qquad {{\textbf{u}}_{i}}\in {{\mathbb{R}}^{L\times 1}},\forall i=1,2,3,...N \\ 
& \qquad{{\textbf{b}}_{i}}\in {{\{-1,1\}}^{L}},\forall i=1,2,3,...N \\ 
\end{aligned}
\end{equation}
where  $\textbf{u}_i$ is an intermediate variable and ${\textbf{b}_i} = sgn({\textbf{u}_i})$.  To optimize the problem in \ref{equ2}, we can optimize the following
regularized problem by moving the equality constraints
in \ref{equ2} to the regularization terms:
\begin{equation}
\label{equ3}
\begin{split}
&\min {{J}}=\frac{1}{2}\sum\limits_{{{s}_{ij}}\in \textbf{S}}{{{\left( \frac{1}{L}{\textbf{u}_{i}}^{T}{\textbf{u}_{j}}-{{s}_{ij}} \right)}^{2}}} 
+  \frac{\lambda_1}{2}\sum\limits_{i = 1}^N {{{\left\| {{\textbf{z}_i} - {\textbf{b}_i}} \right\|}^2}}
+\frac{{{\lambda _2}}}{2}\sum\limits_{i = 1}^N {{{\left\| {{\textbf{b}_i} - {\textbf{u}_i}} \right\|}^2}} \\
& s.t  \quad {\textbf{u}_{i}}\in {{\mathbb{R}}^{L\times 1}},\forall i=1,2,3,...N \\ 
& \qquad{\textbf{b}_{i}}\in {{\{-1,1\}}^{L}},\forall i=1,2,3,...N 
\end{split}
\end{equation}
where $\lambda_{2}$ is the hyper-parameter for the regularization term. Actually, introducing an intermediate variable $\textbf{u}$ is equivalent to adding another full-connected layer between $\textbf{z}$ and $\textbf{b}$ in the hashing layer. To reduce the complexity of our model, we let $\textbf{z}=\textbf{u}$, and then we can have a simplified objective function as:
\begin{equation}
\label{eq.obj2}
\begin{split}
&\min {{J}}=\frac{1}{2}\sum\limits_{{{s}_{ij}}\in \textbf{S}}{{{\left( \frac{1}{K}{\textbf{z}_{i}}^{T}{\textbf{z}_{j}}-{{s}_{ij}} \right)}^{2}}} 
+  \frac{\lambda_1}{2}\sum\limits_{i = 1}^N {{{\left\| {{\textbf{z}_i} - {\textbf{b}_i}} \right\|}^2}} \\
& s.t  \quad{\textbf{b}_{i}}\in {{\{-1,1\}}^{L}},\forall i=1,2,3,...N 
\end{split}
\end{equation}
Eq.~\ref{eq.obj2} is not discrete and $\textbf{z}_i$ is derivable, so we can use back-propagation (BP) to optimize it.

\subsection{Learning}
To learning DDH Model, we need to obtain the parameters of neural networks. We set
\begin{equation}
\label{equ4}
\begin{split}
&{{\textbf{z}}_{i}}={{\mathbf{W}}^{T}}\phi \left( {{x}_{i}};\theta  \right)+\mathbf{c}\\
&{\textbf{b}_{i{\rm{ }}}} = {\mathop{\rm \textit{sgn}}} ({\textbf{z}_i}) = {\mathop{\rm \textit{sgn}}} ({{\bf{W}}^T}\phi \left( {{x_i};\theta } \right) + {\bf{c}})
\end{split}
\end{equation}
where $\theta$ denotes all the parameters of  CNN-F  network for learning the features. $\phi \left( {{x}_{i}};\theta  \right)$ denotes the output of the $full_7$ layer associated with image $x_{i}$. $\mathbf{W}\in {{\mathbb{R}}^{4096\times L}}$ denotes hash layer weights matrix, and $\mathbf{C}\in {{\mathbb{R}}^{L\times 1}}$ is a bias vector. We add regularization terms on the parameters and change the loss function with \ref{eq.obj2} constrains as:
\begin{equation}
\begin{aligned}
\begin{array}{l}
\min {J} = \frac{1}{2}\sum\limits_{{s_{ij}} \in S} {{{\left( {\frac{1}{L}{\Theta _{ij}} - {s_{ij}}} \right)}^2}} \\
+ \frac{{{\lambda _1}}}{2}\sum\limits_{i = 1}^N {{{\left\| {{\textbf{b}_i} - ({{\bf{W}}^T}\phi \left( {{x_i};\theta } \right) + {\bf{c}})} \right\|}^2}} \\
+ \frac{{{\lambda _2}}}{2}{(\left\| {\bf{W}} \right\|_F^2 + \left\| {\bf{c}} \right\|_F^2)^2}
\end{array}
\end{aligned}
\end{equation}
where ${{\Theta }_{ij}}={{z}_{i}}^{T}{{z}_{j}}$, $\lambda_{1}$ and $\lambda_{2}$ are two parameters to balance the effect of different terms.
Stochastic gradient descent (SGD) is used to learn the parameters. We use CNN-F network trained on ImageNet to initialize our network. 
In particular, in each iteration we sample a mini-batch of points from the whole training set and use back-propagation (BP) to optimize the whole network. Here, we compute the derivatives of the loss function as follows:
\begin{equation}
\frac{{\partial {J}}}{{\partial {\textbf{z}_i}}} = \frac{1}{{{L^2}}}({\textbf{z}_i}^T{\textbf{z}_j}){\textbf{z}_j} - {\frac{1}{{{L}}}s_{ij}}{\textbf{z}_j} + {\lambda _1}({\textbf{z}_i} - {\textbf{b}_i})
\end{equation}

\subsection{Out-of-Sample Extension}
After the network has been trained, we still need to obtain the hashing codes of the images which are not in the training data. 
For a novel image, we obtain its binary code by inputing it into the DDH network and make a forward propagation as below:
\begin{equation*}
{\textbf{b}_{i{\rm{ }}}} = {\mathop{\rm \textit{sgn}}} ({\textbf{z}_i}) = {\mathop{\rm \textit{sgn}}} ({{\bf{W}}^T}\phi \left( {{x_i};\theta } \right) + {\bf{c}})
\end{equation*}

\section{Experiment}
Our experiment PC is configured with an Intel(R) Xeon(R) CPU E5-2650 v3 @ 2.30GHz  with 40 cores and the the RAM is 128.0 GB and the GPU is GeForce GTX TITAN X with 12GB.

\subsection{Datasets}
We conduct experiments on three challenging datasets, the Oxford 17 Category Flower Dataset, the CIFAR-10 color images, and the NUS-WIDE. We test our binary descriptor on various tasks,
including  image retrieval and image classification.

\begin{enumerate}
	\item \textbf{CIFAR-10 Dataset} \cite {Krizhevsky2012Learning} contains 10 object categories
	and each class consists of 6,000 images, resulting in a
	total of 60,000 images. The dataset is split into training
	and test sets, with 50,000 and 10,000 images respectively.
	
	\item \textbf{NUS-WIDE dataset}~\cite{Chua2009NUS} has nearly 270,000 images collected from the web. It is a multi-label dataset in which each image is annotated with one or multiple class labels from 81 classes. Following\cite{Lai2015Simultaneous}, we only use the images associated with the 21 most frequent classes. For these classes, the number of images of each class is at least 5000. We use 4,000 for training and 1,000 for testing.
	
	\item \textbf{The Oxford 17 Category Flower Dataset} \cite{nilsback2006visual} contains
	17 categories and each class consists of 80 images, resulting in a total of 1,360 images. The dataset is split into the training (40 images per class), validation (20 images per class), and test (20 images per class) sets.
\end{enumerate}

\subsection{Results on Image Retrieval}
To evaluate the performance of the proposed DDH, we test our method on the task of image
retrieval. We compare DDH with other hashing methods, such as LSH \cite{DBLP:journals/cacm/AndoniI08}, ITQ \cite{DBLP:journals/pami/GongLGP13}, HS \cite{salakhutdinov2009semantic}, Spectral hashing
(SpeH) \cite{NIPS2008_3383}, Spherical hashing (SphH) \cite{heo2012spherical}, KMH \cite{DBLP:conf/cvpr/HeWS13}, Deep Hashing (DH)  \cite{7298862} and DeepBit \cite{Lin_2016_CVPR} ,   Semi-supervised PCAH \cite{wang2010semi} on the CIFAR-10 dataset and NUS-WIDE.
We set the $K_1$=15 and $K_2$=6 to construct labels, and the learning rate as 0.001, $\lambda_{1}$=15, $\lambda_{2}$=0.00001 and batch-size=128. 
Table~\ref{table2} shows the CIFAR-10 retrieval results based on the mean Average Precision (mAP) of the top 1,000 returned images with respect to different bit lengths, while Table~\ref{table4}  shows the mAP value of NUS-WIDE dataset calculated based on the top 5,000 returned neighbors.
The precision/recall in CIFAR-10 dataset is shown in Fig.~\ref{fig.pr}. 

\begin{table}[]
	\centering
	\caption{Performance comparison (mAP) of different unsupervised hashing algorithms on the CIFAR-10 dataset. The mean Average Precision (mAP) are calculated based on the top 1,000 returned images with respect to different number of hash bits.}
	\label{table2}
	\begin{tabular}{l|lllll}		\hline
		Method  &16 bit& 32 bit   &  64 bit\\		\hline \hline
		Method  & 16 bit         & 32 bit         & 64 bit         \\  
		KMH     & 0.136          & 0.139          & 0.145          \\  
		SphH    & 0.145          & 0.146          & 0.154          \\  
		SH      & 0.130          & 0.141          & 0.139          \\  
		PCAH    & 0.129          & 0.126          & 0.121          \\  
		LSH     & 0.126          & 0.138          & 0.157          \\  
		PCA-ITQ & 0.157          & 0.162          & 0.166          \\  
		DH      & 0.162          & 0.166          & 0.170          \\  
		DeepBit &  {0.194}          &  {0.249}          &  {0.277}          \\ \hline
		DDH    & \textbf{0.447} & \textbf{0.486} & \textbf{0.535} \\  
		\hline
	\end{tabular}
\end{table}
\begin{table}[]
	\centering
	\caption{Performance comparison (mAP) of different unsupervised hashing algorithms on the NUS-WIDE dataset. The mAP is calculated based on the top 5,000 returned neighbors for NUS-WIDE dataset.}
	\label{table4}
	\begin{tabular}{l|lllll}		\hline
		Method & 12 bit                              & 24 bit & 32 bit & 48 bit \\  
		\hline		\hline
		CNNH   & 0.611                               & 0.618  & 0.625  & 0.608  \\ 
		FastH  & \textbf{0.621}                               & \textbf{0.650}  & \textbf{0.665}  & \textbf{0.685}  \\ 
		SDH    & 0.568                               & 0.600  & 0.608  & 0.637  \\  
		KSH    & 0.556                               & 0.572  & 0.581  & 0.588  \\  
		LFH    & 0.571                               & 0.568  & 0.568  & 0.585  \\  
		ITQ    &  {0.452}          & 0.468  & 0.472  & 0.477  \\  
		SH     & {0.454}          & 0.406  & 0.405  & 0.400  \\  \hline
		DDH   &  {\textbf{0.675}} & \textbf{0.680}       & \textbf{0.701}       &   \textbf{0.712}    \\ 
		\hline
	\end{tabular}
\end{table}

From these results, we have the following observations:

\noindent 1) Our method significantly outperforms the other deep or non-deep hashing methods in all datasets. In CIFAR-10, the improvement of DDH over the other methods is more significant, compared with that in NUS-WIDE dataset. Specifically, it outperforms the best counterpart (DeepBit) by 25.3\%, 23.7\% and 25.8\% for 16, 32 and 64-bit hash codes. One possible reason is that CIFAR-10 contains simple images, and the constructed neighborhood structure is more accurate than the other two datasets. DDH improves the state-of-the-arts by 5.4\%, 3.0\%, 3.6\% and 2.7\% in NUS-WIDE dataset.

\noindent 2) Table~\ref{table2} shows that DeepBit and FashH are strong competitors in terms of mAP in CIFAR-10 and NUS-WIDE dataset. But the performance gap of DeepBit and our DDH is still very large, which is probably due to that DeepBit uses only 3 fully connected layers to extract the features.
Fig.~\ref{fig.pr} shows that most of the hashing methods can achieve a high recall for small number of retrieved samples (or recall). But obviously, our DDH significantly outperforms the others.

\noindent 3) With the increase of code length, the performance of most hashing methods is improved accordingly. An exception is PCAH, which has no improvement with the increase of code length.

\begin{figure}[h]
	\centering
	\subfigure[{16 bits}]{
		\includegraphics[width=0.30\linewidth]{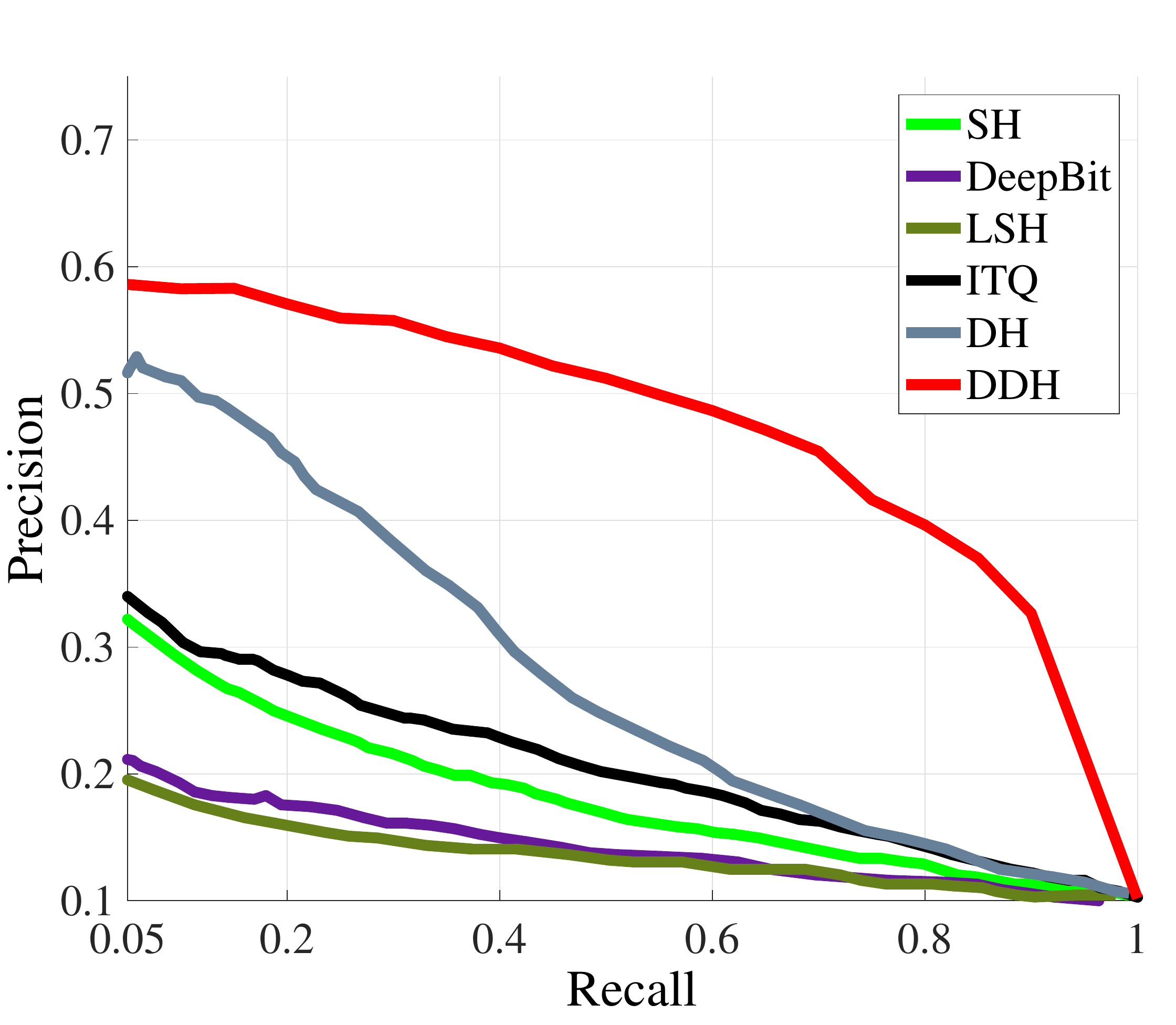}}
	\subfigure[{32 bits}]{
		\includegraphics[width=0.30\linewidth]{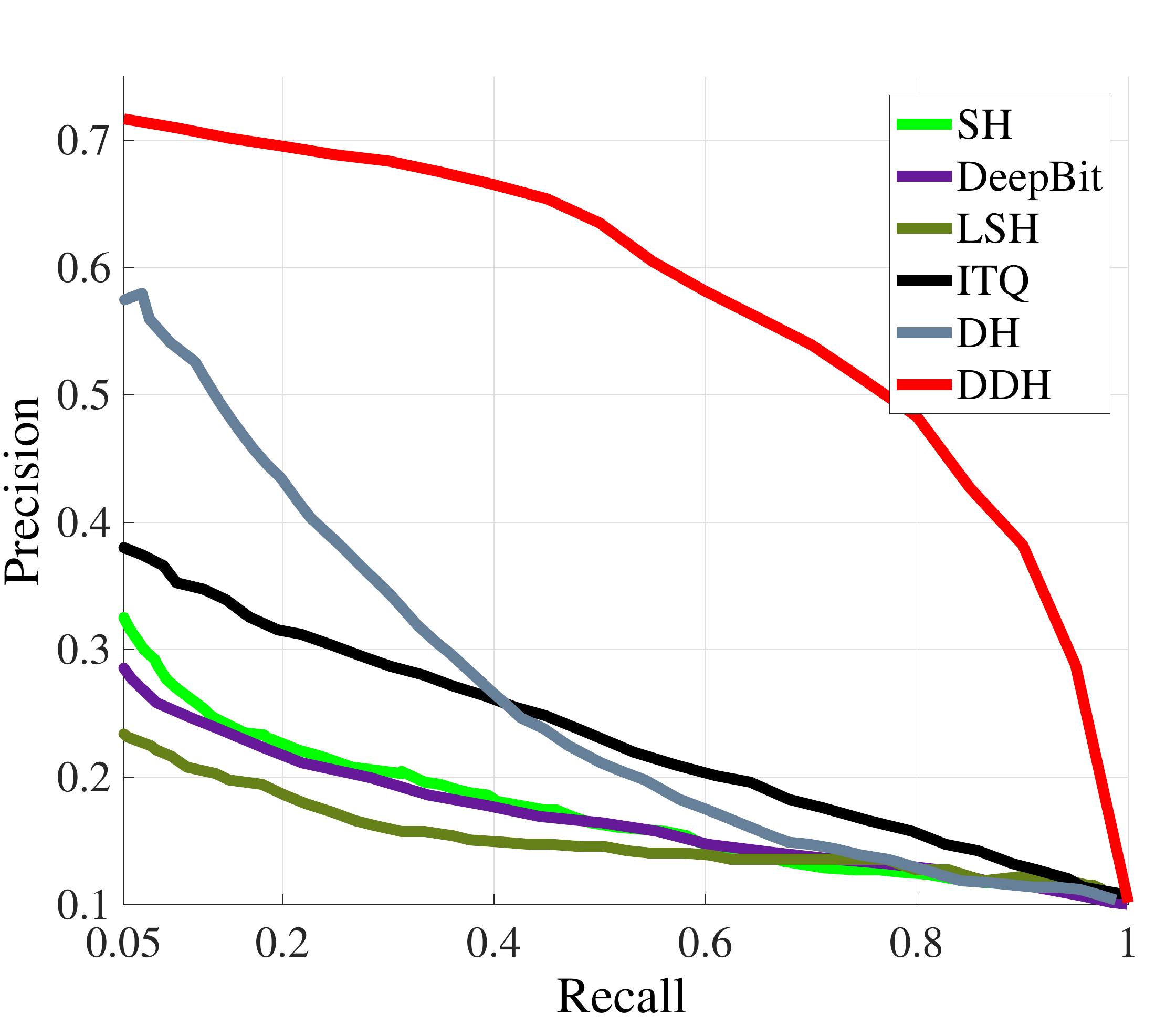}}
	\subfigure[{64 bits}]{
		\includegraphics[width=0.30\linewidth]{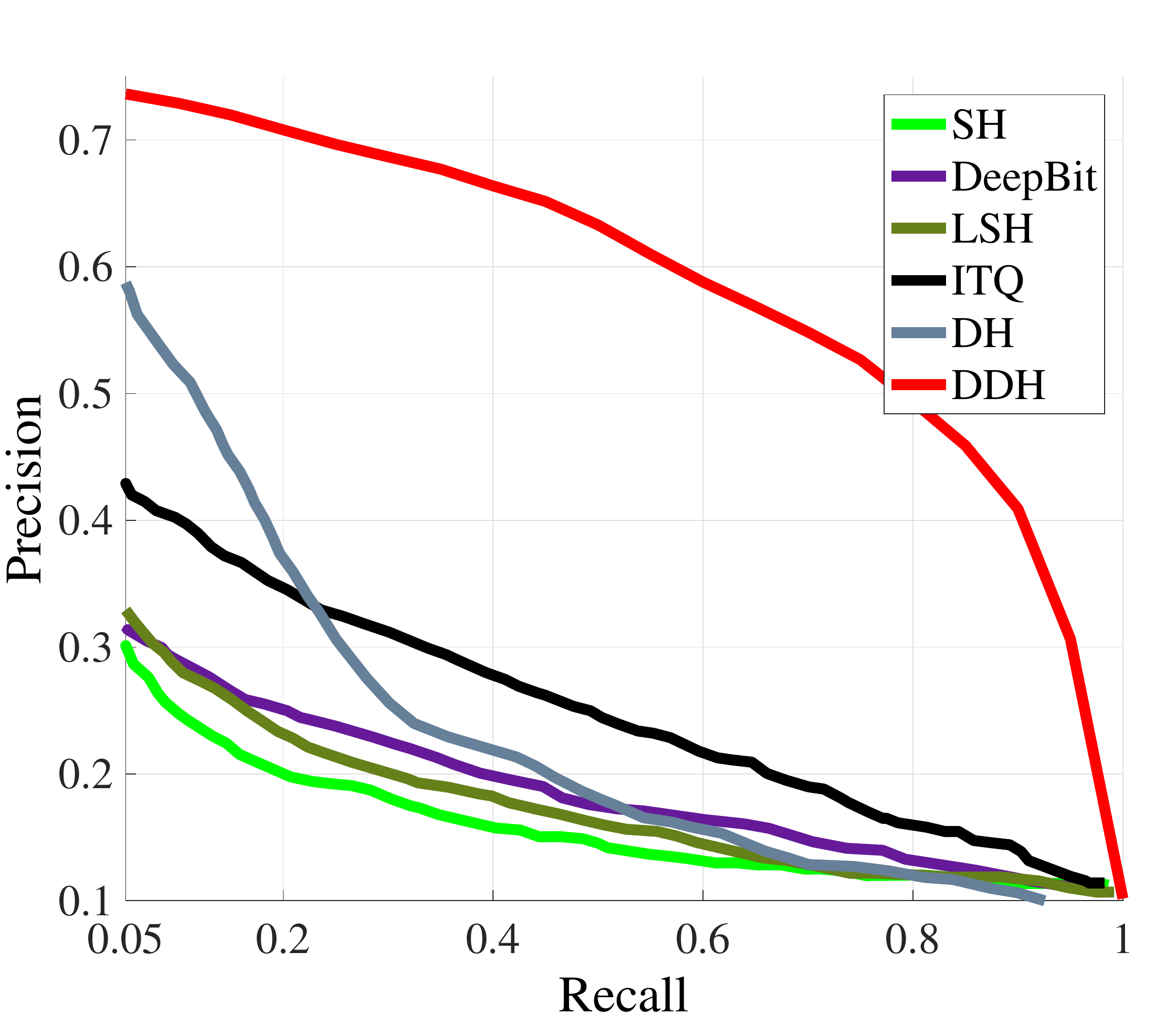}}
	\caption{Precision/Recall curves of different unsupervised hashing methods on the CIFAR-10 dataset with respect to 16, 32 and 64 bits, respectively}
	\label{fig.pr}
\end{figure}
To make fair comparison with the non-deep hashing methods, and validate that our improvement is not only caused by the deep features, we conduct non-deep hashing methods with deep features extracted by the CNN-F pre-trained on ImageNet.
The results are reported in Table~\ref{table3}, where ``ITQ+CNN'' denotes the ITQ method with
deep features and other methods have similar notations.
When we run the non-deep hashing methods on deep features, the performance is usually improved compared with the hand-crafted features. 

\begin{table}[]
	
	\centering
	\caption{Performance comparison (mAP) of different hashing algorithms with deep features on the CIFAR-10 dataset.}
	\label{table3}
	\begin{tabular}{l|lllll}
		\hline
		Method  & 12 bit         & 24 bit         & 32 bit    &48 bit     \\  
		\hline\hline

		ITQ + CNN     & 0.237          & 0.246          & 0.255  &0.261       \\ 
		SH + CNN      &0.183         & 0.164          & 0.161       &0.161   \\  
		
		SPLH + CNN     & \textbf{0.299}          & \textbf{0.330}          & \textbf{0.335}   &\textbf{0.330}      \\  
		LFH + CNN    & 0.208         & 0.242          & 0.266    &0.339     \\  \hline
		DDH    & \textbf{0.414} & \textbf{0.467} & \textbf{ 0.486 } &\textbf{0.512} \\ 
		\hline
	\end{tabular}
\end{table}

 By constructing the neighborhood structure using the labels, our method can be easily modified as a supervised hashing method.
 Therefore, we also compared with supervised hashing methods, and show the mAP results on NUS-WIDE dataset in Table~\ref{tabel5}. It is obvious that our DDH outperforms the state-of-the-art deep and non-deep supervised hashing algorithms by a large margin, which are 5.7\%, 5.8\%, 7.8\% and 8.1\% for 12, 24, 32, and 48-bits hash codes. This indicates that the performance improvement of DDH is not only due to the constructed neighborhood structure, but also the other components.
 \begin{table}[h]
 	\centering
 	\caption{Results on NUS-WIDE. The table shows other deep network with supervised pair-wise labels. The mAP value is calculated	based on the top 5000 returned neighbors for NUS-WIDE dataset.}
 	\label{tabel5}
 	\begin{tabular}{l|lllll}
 		\hline
 		Method & 16 bit                              & 24 bit & 32 bit & 48 bit \\ 
 		\hline \hline
 		\noalign{\smallskip}
 		DRSCH  & \textbf{0.618}                               & \textbf{0.622}  & \textbf{0.623}  & 0.628  \\  
 		DSCH   & 0.592                               & 0.597  & 0.611  & 0.609  \\ 
 		DSRH   & 0.609                               & 0.618  & 0.621  & \textbf{0.631}  \\  \hline
 		DDH   & \textbf{0.675}  & \textbf{0.680}       & \textbf{0.701}       & \textbf{0.712}    \\ 
 		\hline
 	\end{tabular}
 \end{table}

\subsection{Results on Object Recognition}

In the task of object recognition, the algorithm needs to recognize very similar object (daisy, iris and pansy etc.).   So it requires more discriminative binary codes to represent images that look very similar. In this paper, we use the Oxford 17 Category Flower Dataset to evaluate our method on object recognition and we compared with several real-valued descriptors such as HOG \cite{dalal2005histograms} and SIFT.

Due to the variation of color distributions, pose deformations and shapes, ``Flower'' recognition becomes more challenging. Besides, we need to consider the computation cost while one wants to recognize the flowers in the wild using mobile devices, which makes us generate very short and efficient binary codes to discriminate flowers.
Following the setting in \cite{nilsback2006visual}, we train a multi-class SVM classifier with our proposed binary descriptor. Table \ref{table:flower} compares the classification accuracy of the 17 categories flowers using different descriptors proposed in \cite{nilsback2006visual}, \cite{nilsback2008automated}, including low-level (Color, Shape, Texture), and high-level (SIFT and HOG) features. Our proposed binary descriptor with 256 dimensionality improves previous best recognition accuracy by around 5.01\% (80.11\% vs. 75.1\%). We also test our proposed method with 64 bits, which still outperforms the state-of-art result (76.35\% vs. 75.1\%). We also test the computational complexity during SVM classifier training with only costing 0.3s training on 256 bits and 0.17s  on 64 bits. Compared with other descriptors, such as Color, Shape, Texture, HOG, HSV and SIFT, DDH demonstrates its efficiency and effectiveness. \begin{table}[h]
	\centering
	\caption{The categorization accuracy (mean\%) and training time for
		different features on the Oxford 17 Category Flower
		Dataset}
	\label{table:flower}
	\begin{tabular}{llllll}
		\hline\noalign{\smallskip}
		Descriptors   & Accuracy       & Time       \\  
		\noalign{\smallskip}
		\hline
		\noalign{\smallskip}
		Colour        & 60.9           & 3          \\  
		Shape         & 70.2           & 4          \\  
		Texture       & 63.7           & 3          \\  
		HOG           & 58.5           & 4          \\  
		HSV           & 61.3           & 3          \\  
		SIFT-Boundary & 59.4           & 5          \\ 
		SIFT-Internal & 70.6           & 4          \\  
		DeepBit (256bits)  & \textbf{75.1}           & \textbf{0.07}       \\  
		DDH (64bits)      & 76.4 & 0.12
		\\  
		DDH(256bits)     & \textbf{80.5} &  \textbf{0.30}          \\ \hline
	\end{tabular}
\end{table}
 
 \section{Conclusion and Future work}
 \label{sec.conclusion}
 In this work, we address two central problems remaining largely unsolved for image hashing: 1) how to directly generate binary codes without relaxation? 2) how to equip the binary representation with the ability of accurate image retrieval? 
 We resolve these problems by introducing an intermediate variable and a loss function steering the learning process, which is based on the neighborhood structure in the original space.
 Experiments on real datasets show that our method can outperform other unsupervised and supervised methods to achieve the state-of-the-art performance in image retrieval and object recognition.
 In the future, it is necessary to improve the classification accuracy by incorporating a classification layer at the end of this architecture.
 
 \section{Acknowledgment}
 
 This work was supported in part by the National Natural Science Foundation of China under Project 61502080, Project 61632007, 
 and the Fundamental Research Funds for the Central Universities under Project ZYGX2016J085, Project ZYGX2014Z007.

\end{document}